\documentclass{article} 
\usepackage{nips13submit_e,times}

\usepackage{times}
\usepackage{epsfig}
\usepackage{graphicx}
\usepackage{amsmath}
\usepackage{amssymb}
\usepackage{url}
\newcommand{\Conv}{\mathop{\scalebox{1.5}{\raisebox{-0.2ex}{$\ast$}}}}%

\usepackage[center]{subfigure}
\usepackage{tabularx}
\usepackage{booktabs}

\usepackage[pagebackref=true,breaklinks=true,letterpaper=true,colorlinks,bookmarks=false]{hyperref}

\nipsfinalcopy

\usepackage{textcomp} 

\begin{document}


\title{Learning Human Pose Estimation Features with Convolutional Networks}

\author{Arjun Jain\\
New York University\\
{\tt\small ajain@nyu.edu}
\and
Jonathan Tompson\\
New York University\\
{\tt\small tompson@cims.nyu.edu}
\and
Mykhaylo Andriluka\\
MPI Saarbruecken\\
{\tt\small andriluk@mpi-inf.mpg.de}
\and
Graham W. Taylor\\
University of Guelph\\
{\tt\small gwtaylor@uoguelph.ca}
\and
Christoph Bregler\\
New York University\\
{\tt\small chris.bregler@nyu.edu}
}

\maketitle

\begin{abstract}
This paper introduces a new architecture for human pose estimation using a multi-layer convolutional network architecture and a modified learning technique that learns low-level features and a higher-level weak spatial model.
Unconstrained human pose estimation is one of the hardest problems in computer vision, and our new architecture and learning schema shows improvement over the current state-of-the-art.
The main contribution of this paper is showing, for the first time, that a specific variation of deep learning is able to meet the performance, and in many cases outperform, existing traditional architectures on this task.
The paper also discusses several lessons learned while researching alternatives, most notably, that it is possible to learn strong low-level feature detectors on regions that might only cover a few pixels in the image.
Higher-level spatial models improve somewhat the overall result, but to a much lesser extent than expected.
Many researchers previously argued that the kinematic structure and top-down information are crucial for this domain, but with our purely bottom-up, and weak spatial model, we improve on other more complicated architectures that currently produce the best results.
This echos what many other researchers, like those in the speech recognition, object recognition, and other domains have experienced \cite{lucchi2011spatial}.

\end{abstract}

\begin{figure}[ht]
\begin{center}
\includegraphics[width=1\linewidth]{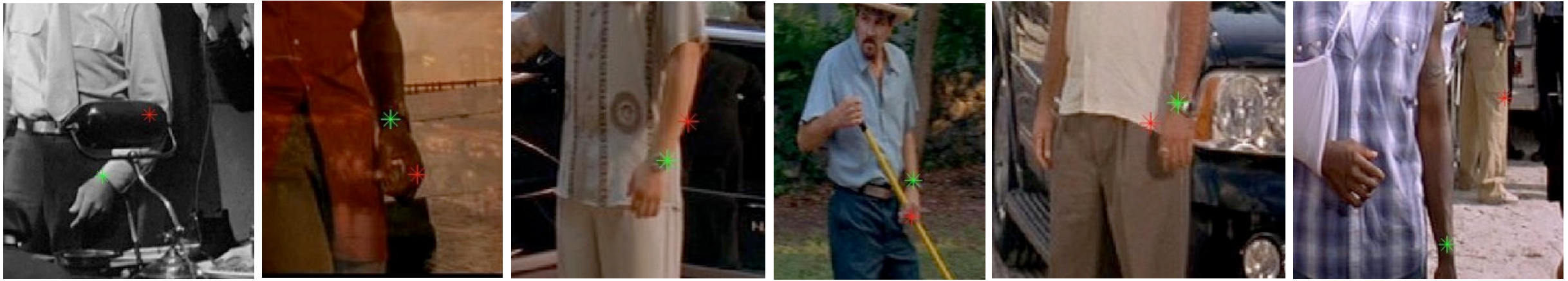}
\end{center}
\label{fig:teaser}
\caption{The green cross is our new technique's wrist locator, the red cross is the state-of-the-art CVPR13 MODEC detector \cite{sapp13cvpr} on the FLIC database.}
\end{figure}

\section{Introduction}

One of the hardest tasks in computer vision is determining the high
degree-of-freedom configuration of a human body with all its limbs,
complex self-occlusion, self-similar parts, and large variations due to clothing,
body-type, lighting, and many other factors.  The most challenging scenario for
this problem is from a monocular RGB image and with no prior assumptions
made using motion models, pose models, background models, or any other
common heuristics that current state-of-the-art systems
utilize. Finding a face in frontal or side view is relatively simple,
but determining the exact location of body parts such as hands,
elbows, shoulders, hips, knees and feet, each of which sometimes only
occupy a few pixels in the image in front of an arbitrary cluttered
background, is significantly harder.

The best performing pose estimation methods, including those based on
deformable part models, typically are based on body part detectors.
Such body part detectors commonly consist of multiple stages of processing.
The first stage
of processing in a typical pipeline consists of extracting sets of
low-level features such as SIFT \cite{lowe1999object}, HoG
\cite{dalal2005histograms}, or other filters that describe orientation statistics in
local image patches. Next, these features are pooled over local spatial
regions and sometimes across multiple scales to reduce the size of the
representation and also develop local shift/scale invariance. Finally,
the aggregate features are mapped to a vector, which is then either
input to 1) a standard classifier such as a support vector machine
(SVM) or 2) the next stage of processing (e.g.~assembling the parts
into a whole). Much work is devoted to engineering the system to
produce a vector representation that is sensitive to class (e.g.~head,
hands, torso) while remaining invariant to the various nuisance
factors (lighting, viewpoint, scale, etc.)

An alternative approach is \emph{representation learning}: relying on the
data instead of feature engineering, to \emph{learn} a good
representation that is invariant to nuisance
factors. For a recent review, see \cite{Bengio2012}. It is common
to learn multiple layers of representation, which is referred to as
\emph{deep learning}. Several such techniques have used unsupervised or
semi-supervised learning to extract multi-layer domain-specific
invariant representations, however, it is purely supervised techniques
that have won several recent challenges by large margins, including ImageNet 
LSVRC 2012 and 2013~\cite{krizhevsky2012imagenet,zeiler2013visualizing}. These end-to-end learning systems have
capitalized on advances in computing hardware (notably GPUs), larger
datasets like ImageNet, and algorithmic advances (specifically
gradient-based training methods and regularization).

While these methods are now proven in generic object recognition,
their use in pose estimation has been limited. Part of the challenge
in making end-to-end learning work for human pose estimation is
related to the nonrigid structure of the body, the necessity for
precision (deep recognition systems often throw away precise location
information through pooling), and the complex, multi-modal nature of
pose.

In this paper, we present the first end-to-end learning approach for
full-body human pose estimation. While our approach is based on
convolutional networks (convnets) \cite{LeCun1998}, we want to
stress that the na\"ive implementation of applying this model
``off-the-shelf'' will not work. Therefore, the contribution of this
work is in both a model that outperforms state of the art deformable
part models (DPMs) on a modern, challenging dataset, and also
an analysis of what is needed to make convnets work in human
pose estimation. In particular, we present a two-stage filtering
approach whereby the response maps of convnet part detectors are
denoised by a second process informed by the part hierarchy.
\label{sec:intro}


\section{Related Work}
\label{sec:relatedwork}

Detecting people and their pose has been investigated for decades.
Many early techniques rely on sliding-window part detectors based on
hand-crafted or learned features or silhouette extraction techniques
applied to controlled recording conditions.  Examples include
\cite{farhadi2007transfer,wren1997pfinder,athitsos2004boostmap,nowlan1995convolutional}.
We refer to \cite{poppe2007vision} for a complete survey of this era.
More recently, several new approaches have been proposed that
are applied to unconstrained domains.  In such domains, good performance has
been achieved with so-called ``bag of features'' followed by regression-based,
nearest neighbor or SVM-based architectures.  Examples include ``shape-context''
edge-based histograms from the human body
\cite{mori2002estimating,agarwal2006recovering} or just silhouette
features \cite{Grauman2003}.  Shakhnarovich et al. \cite{Shakhnarovich2003}
learn a parameter sensitive hash function to perform example-based pose estimation.  
Many relevant techniques have also been applied to hand tracking
such as \cite{wang2009real}.
A more general survey of the large field of hand tracking can be found
in \cite{erol2007vision}.

Many techniques have
been proposed that extract, learn, or reason over entire body
features.  Some use a combination of local detectors and structural
reasoning (see \cite{ramanan2005strike} for coarse tracking and
\cite{buehler2009learning} for person-dependent tracking).  In a
similar spirit, more general techniques using pictorial structures
\cite{andriluka2009pictorial,andriluka2010monocular,Ferrari2009,Sapp2010,pishchulin12cvpr,pishchulin11bmvc}, ``poselets''
\cite{PoseletsICCV09}, and other part-models
\cite{Felzenszwalb2010PAMI,yang2011articulated} have received increased
attention.  We will focus on these techniques and their latest incarnations
in the following sections.

Further examples come from the HumanEva dataset competitions
\cite{Sigal2010}, or approaches that use higher-resolution shape models such as SCAPE \cite{anguelov2005scape}
and further extensions \cite{HasStoSunRosSei09,Coreg:Patent:2012}.
These differ from our domain in that the images considered are of
higher quality and less cluttered.
Also many of these techniques work on images from a single camera, but need video
sequence input (not single images) to achieve impressive results
\cite{stoll2011fast,zuffiestimating}.

As an example of a technique that works for single images against cluttered backgrounds,
Shotton et al.'s~Kinect based body part detector
\cite{shotton2013real} uses a random forest of decision trees trained
on synthetic depth data to create simple body part detectors. In the
proposed work, we also adopt simple part-based detectors, however, we
focus on a different learning strategy.

There are a number of successful end-to-end
representation learning  techniques which perform pose estimation
on a limited subset of body parts or body poses. One of the earliest examples of
this type was Nowlan and Platt's
convolutional neural network hand tracker
\cite{nowlan1995convolutional}, which tracked a single hand. Osadchy
et al.~applied a convolutional network to simultaneously detect and
estimate the pitch, yaw and roll of a face
\cite{osadchy2007synergistic}. Taylor et al.~\cite{taylor2010embedding} trained a
convolutional neural network to learn an embedding in which images of
people in similar pose lie nearby. They used a subset of body parts,
namely, the head and hand locations to learn the ``gist'' of a pose,
and resorted to nearest-neighbour matching rather than explicitly
modeling pose. Perhaps most relevant to our work is Taylor et al.'s work on
tracking people in video \cite{taylor2010tracking}, augmenting a particle filter with a
structured prior over human pose and dynamics based on learning
representations. While they estimated a posterior over the whole body
(60 joint angles), their experiments were limited to the HumanEva
dataset \cite{Sigal2010},
which was collected in a controlled laboratory setting. The datasets
we consider in our experiments are truly poses ``in the wild'', though
we do not consider dynamics.

A factor limiting earlier methods from tacking full pose-estimation
with end-to-end learning methods, in particular deep networks, was the
limited amount of labeled data. Such techniques, with millions or more
parameters, require more data than structured techniques that have
more \emph{a priori} knowledge, such as DPMs. We attack this issue on
two fronts. First, directly, by using larger labeled training sets
which have become available in the past year or two, such as FLIC
\cite{sapp13cvpr}. Second, indirectly, by better exploiting the data
we have. The annotations provided by typical pose estimation datasets
contain much richer information compared to the class labels in object
recognition datasets In particular, we show that the relationships
among parts contained in these annotations can be used to build better
detectors.


\section {Model}
\label{sec:model}

To perform pose estimation with a convolutional network architecture~\cite{LeCun1998} (convnet), the most obvious approach would be to map the image input directly to a vector coding the articulated pose: i.e. the type of labels found in pose datasets. The convnet output would represent the unbounded 2-D or 3-D positions of joints, or alternatively a hierarchy of joint angles. However, we found that this worked very poorly. One issue is that pooling, while useful for improving translation invariance during object recognition, destroys precise spatial information which is necessary to accurately predict pose. Convnets that produce segmentation maps, for example, avoid pooling completely \cite{Turaga2010, FarabetCouprieNajmanLeCun2012}. Another issue is that the direct mapping from input space to kinematic body pose coefficients is highly non-linear and not one-to-one.  However, even if we took this route, there is a deeper issue with attempting to map directly to a representation of full body pose. Valid poses represent a much lower-dimensional manifold in the high-dimensional space in which they are captured. It seems troublesome to make a discriminative network map to a space in which the majority of configurations do not represent valid poses. In other words, it makes sense to restrict the net's output to a much smaller class of valid configurations.


Rather than perform multiple-output regression using a single convnet to learn pose coefficients directly, we found that training multiple convnets to perform independent binary body-part classification, with one network per feature, resulted in improved performance on our dataset.  These convnets are applied as \emph{sliding windows} to overlapping regions of the input, and map a window of pixels to a single binary output: the presence or absence of that body part. The result of applying the convnet is a \emph{response-map} indicating the confidence of the body part at that location. This lets us use much smaller convnets, and retain the advantages of pooling, at the expense of having to maintain a separate set of parameters for each body part. Of course, a series of independent part detectors cannot enforce consistency in pose in the same way as a structured output model, which produces valid full-body configurations. In the following sections, we first describe in detail the convolutional network architecture and then a method of enforcing pose consistency using parent-child relationships.
\subsection{Convolutional Network Architecture}
\label{sec:convnet}

\begin{figure}
        \centering
        \includegraphics[width=\textwidth]{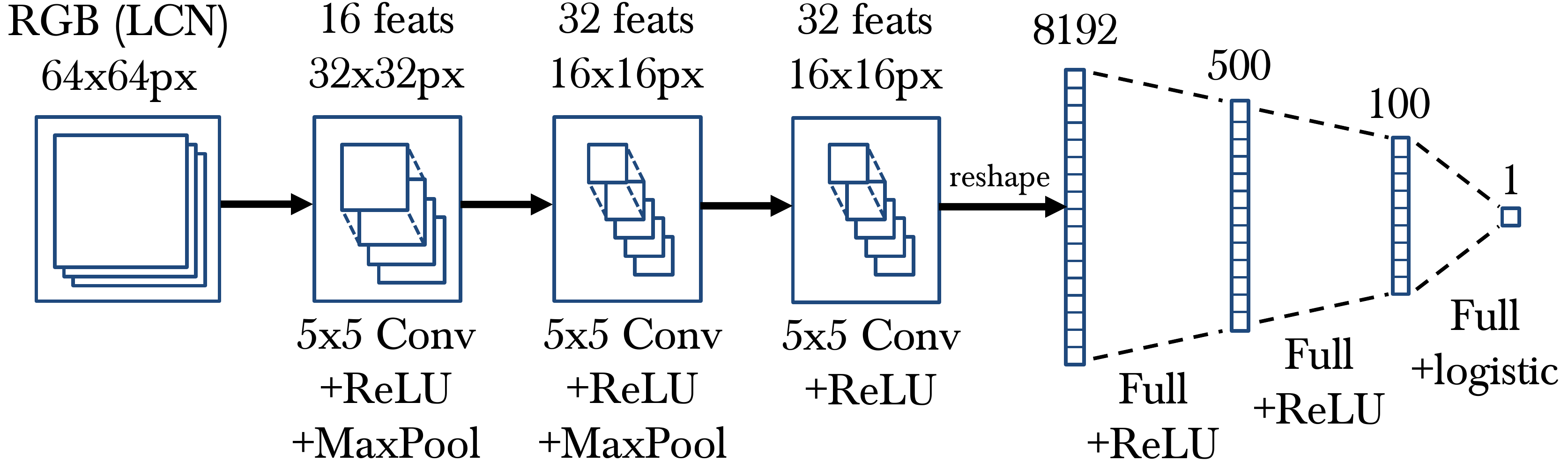}
        \caption{The convolutional network architecture used in our experiments.}
        \label{fig:network_conv}
\end{figure}

The lowest level of our two-stage feature detection pipeline is based on a standard convnet architecture, an overview of which is shown in Figure \ref{fig:network_conv}. Convnets, like their fully-connected, deep neural network counterparts, perform end-to-end feature learning and are trained with the back-propagation algorithm. However, they differ in a number of respects, most notably local connectivity, weight sharing, and local pooling. The first two properties significantly reduce the number of free parameters, and reduce the need to learn repeated feature detectors at different locations of the input. The third property makes the learned representation invariant to small translations of the input.

The convnet pipeline shown in Figure \ref{fig:network_conv} starts with a 64$\times$64 pixel RGB input patch which has been local contrast normalized (LCN)~\cite{yann_lcn_cite} to emphasize geometric discontinuities and improve generalization performance \cite{pinto2008real}. The LCN layer is comprised of a 9$\times$9 pixel local subtractive normalization, followed by a 9$\times$9 local divisive normalization.  The input is then processed by three convolution and subsampling layers, which use rectified linear units (ReLUs)~\cite{glorot2011deep} and max-pooling.


As expected, we found that internal pooling layers help to a) reduce computational complexity\footnote{The number of operations required to calculate the output of the the three fully-connected layers is $O\left(n^2\right)$ in the size of the $\mathbb{R}^n$ input vectors. Therefore, even small amounts of pooling in earlier stages can drastically reduce training time.} and b) improve classification tolerance to small input image translations. Unfortunately, pooling also results in a loss of spatial precision.  Since the target application for this convnet was offline (rather than real-time) body-pose detection, and since we found that with sufficient training exemplars, invariance to input translations can be learned, we choose to use only 2 stages of $2\times2$ pooling (where the total image downsampling rate is $4\times4$).

Following the three stages of convolution and subsampling, the top-level pooled map is flattened to a vector and processed by three \emph{fully connected} layers, analogous to those used in deep neural networks. Each of these output stages is composed of a linear matrix-vector multiplication with learned bias, followed by a point-wise non-linearity (ReLU). The output layer has a single logistic unit, representing the probability of the body part being present in that patch.


To train the convnet, we performed standard batch stochastic gradient descent.  From the training set images, we set aside a validation set to tune the network hyper-parameters, such as number and size of features, learning rate, momentum coefficient, etc. We used Nesterov momentum \cite{sutskeverimportance} as well as RMSPROP \cite{tieleman2012rmsprop} to accelerate learning and we used L2 regularization and dropout \cite{hinton2012improving} on the input to each of the fully-connected linear stages to reduce over-fitting the restricted-size training set.

\subsection{Enforcing Global Pose Consistency with a Spatial Model}

When applied to the validation set, the raw output of the network
presented in Section~\ref{sec:convnet} produces many
false-positives. We believe this is due to two factors: 1) the small
image context as input to the convnet (64$\times$64 pixels or
approximately 5\% of the input image area) does not give the model
enough contextual information to perform anatomically consistent joint
position inference and 2) the training set size is limited. We
therefore use a higher-level spatial model with simple body-pose
priors to remove strong outliers from the convnet output.  We do not
expect this model to improve the performance of poses that are close
to the ground truth labels (within 10 pixels for instance), but rather it functions as a post processing step to de-emphasize anatomically impossible poses due to strong outliers.

\begin{figure}
        \centering
        \includegraphics[width=0.6\textwidth]{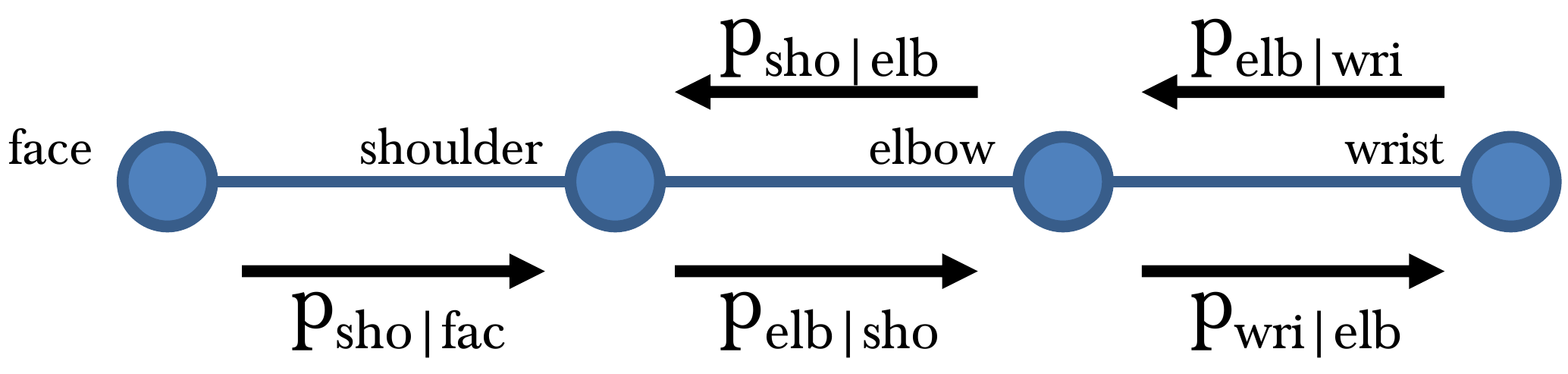}
        \caption{Spatial Model Connectivity with Spatial Priors}
        \label{fig:spacmodel}
\end{figure}

The inter-node connectivity of our simple spatial model is displayed in Figure~\ref{fig:spacmodel}.  It consists of a linear chain of kinematic 2D nodes for a single side of the human body.  Throughout our experiments we used the left shoulder, elbow and wrist; however we could have used the right side joints without loss of generality (since detection of the right body parts simply requires a horizontal mirror of the input image). For each node in the chain, our convnet detector generates response-map unary distributions $p_{\text{fac}}\left(x\right)$, $p_{\text{sho}}\left(x\right)$, $p_{\text{elb}}\left(x\right)$, $p_{\text{wri}}\left(x\right)$ over the dense pixel positions $x$, for the face, shoulder, elbow and wrist joints respectively. For the remainder of this section, all distributions are assumed to be a function over the pixel position, and so the $x$ notation will be dropped. The output of our spatial model will produce filtered response maps: $\hat{p}_{\text{fac}}$, $\hat{p}_{\text{sho}}$, $\hat{p}_{\text{elb}}$, and $\hat{p}_{\text{wri}}$.

\begin{figure}[h]
    \begin{tabular}{ccc}
        \includegraphics[width=0.3\linewidth]{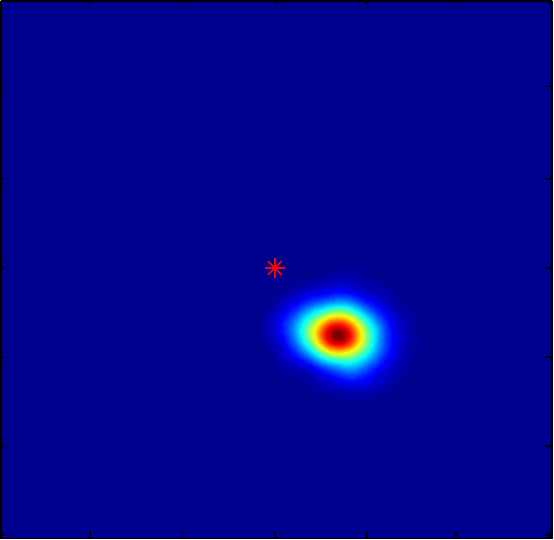} &
        \includegraphics[width=0.3\linewidth]{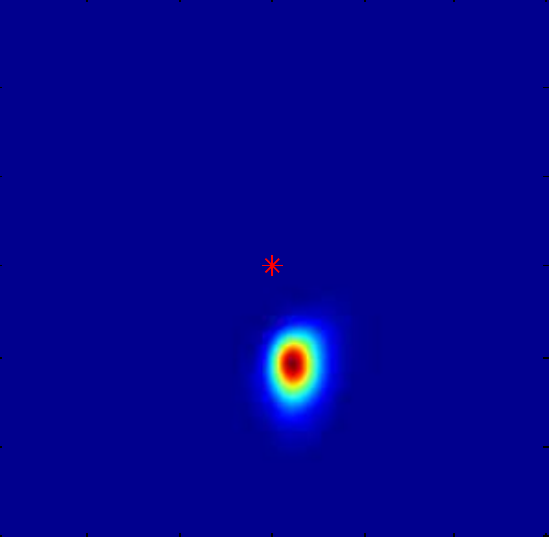} &
        \includegraphics[width=0.3\linewidth]{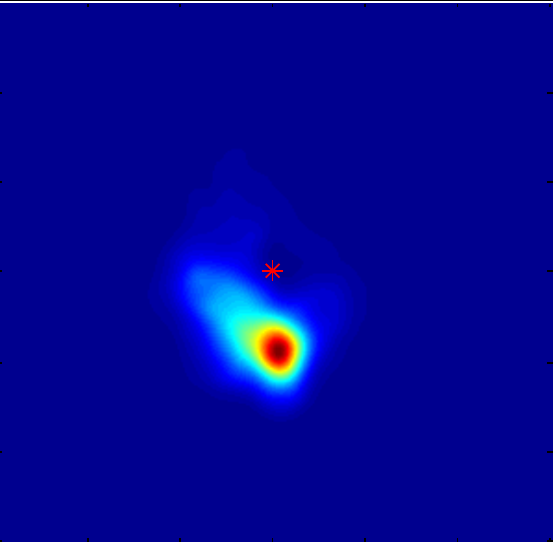} \\
        a) $p_{\text{sho}|\text{fac}=\vec{0}}$ & b) $p_{\text{elb}|\text{sho}=\vec{0}}$ &  c) $p_{\text{wri}|\text{elb}=\vec{0}}$ \\
    \end{tabular}
    \caption{Part priors for left body parts}
    \label{fig:histparts}
\end{figure}

The body part priors for a pair of joints $\left(a,b\right)$, $p_{a|b=\vec{0}}$, are calculated by creating a histogram of joint $a$ locations over the training set, given that the adjacent joint $b$ is located at the image center ($x=\vec{0}$).  The histograms are then smoothed (using a gaussian filter) and normalized. The learned priors for $p_{\text{sho}|\text{fac}=\vec{0}}$, $p_{\text{elb}|\text{sho}=\vec{0}}$, and $p_{\text{wri}|\text{elb}=\vec{0}}$ are shown in Figure~\ref{fig:histparts}. Note that due to symmetry, the prior for $p_{\text{elb}|\text{wri}=\vec{0}}$ is a 180\textdegree~rotation of $p_{\text{wri}|\text{elb}=\vec{0}}$ (as is the case of other adjacent pairs).  Rather than assume a simple Gaussian distribution for modeling pairwise interactions of adjacent nodes, as is standard in many parts-based detector implementations, we have found that the these non-parametric spatial priors lead to improved detection performance.

Given the full set of prior conditional distributions and the convnet unary distributions, we can now construct the filtered distribution for each part by using an approach that is analogous to the sum-product belief propagation algorithm. For body part $i$, with a set of neighbouring nodes $U$, the final distribution is defined as:

\begin{equation}
\hat{p}_i \propto {p_i}^\lambda \text{ } \prod_{u\in
  U}\left(p_{i|u=\vec{0}} \Conv p_u\right)
\end{equation}

where $\lambda$ is a mixing parameter and controls the confidence of each joint's unary distribution towards its final filtered distribution (we used $\lambda=1$ for our experiments). The final joint distribution is therefore a product of the unary distribution for that joint, as well as the beliefs from neighbouring nodes (as with standard sum-product belief propagation).  In log space, the above product for the shoulder joint becomes:

\begin{equation}
\operatorname{log}\left(\hat{p}_{\text{sho}}\right) \propto
\lambda\text{ }\operatorname{log}\left(p_{\text{sho}}\right) +
\operatorname{log}\left({p_{\text{sho}|\text{fac}=\vec{0}}} \Conv
  p_{\text{fac}}\right) +
\operatorname{log}\left({p_{\text{sho}|\text{elb}=\vec{0}}} \Conv
  p_{\text{elb}}\right)
\end{equation}

\begin{figure}
    \centering
    \includegraphics[width=0.3\linewidth]{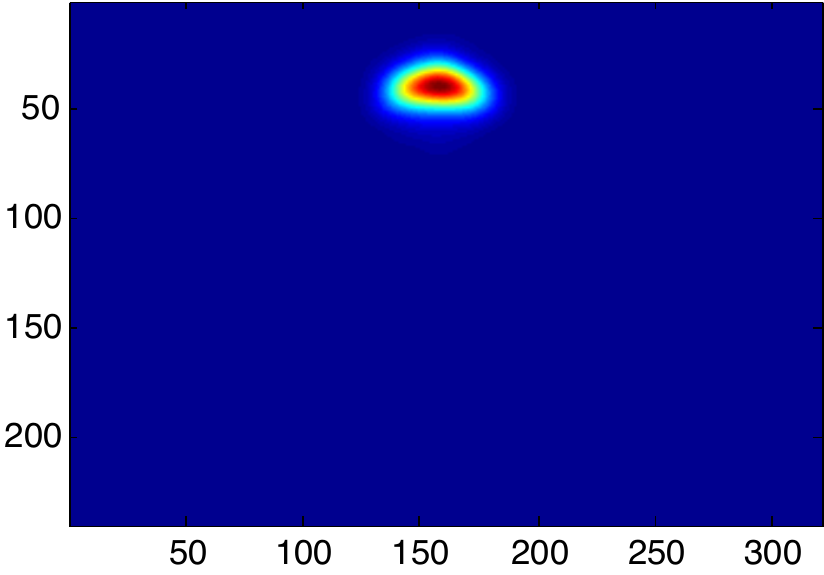}
    \caption{Global prior for the face: $h_{\text{fac}}$}
    \label{fig:facfilter}
\end{figure}

We also perform an equivalent computation for the elbow and wrist joints.  The face joint is treated as a special case.  Empirically, we found that incorporating image evidence from the shoulder joint to the filtered face distribution resulted in poor performance.  This is likely due to the fact that the convnet does a very good job of localizing the face position, and so incorporating noisy evidence from the shoulder detector actually increases uncertainty.  Instead, we use a global position prior for the face, $h_{\text{fac}}$, which is obtained by learning a location histogram over the face positions in the training set images, as shown in Figure~\ref{fig:facfilter}. In log space, the output distribution for the face is then given by:

 \begin{equation}
 \operatorname{log}\left(\hat{p}_{\text{fac}}\right) \propto
 \lambda\text{ }\operatorname{log}\left(p_{\text{fac}}\right) +
 \operatorname{log}\left(h_{\text{fac}}\right)
 \end{equation}


Lastly, since the learned neural network convolution features and the spatial priors are not explicitly invariant to scale, we must run the convnet and spatial model on images at multiple scales at test time, and then use the most likely joint location across those scales as the final joint location.  For datasets containing examples with multiple persons (known a priori), we use non-maximal suppression~\cite{Neubeck:2006:ENS:1170749.1172615} to find multiple local maxima across the filtered response-maps from each scale, and we then take the top $n$ most likely joint candidates from each person in the scene.


\section{Results}
\label{sec:results}

We evaluated our architecture on the FLIC~\cite{sapp13cvpr} dataset, which is comprised of 5003 still RGB images taken from an assortment of Hollywood movies. Each frame in the dataset contains at least one person in a frontal pose (facing the camera), and each frame was processed by Amazon Mechanical Turk to obtain ground truth labels for the joint positions of the upper body of a single person.  The FLIC dataset is very challenging for state-of-the-art pose estimation methodologies because the poses are unconstrained, body parts are often occluded, and clothing and background are not consistent.

We use $3987$ training images from the dataset, which we also mirror horizontally to obtain a total of $3987 \times 2 = 7974$ examples. Since the training images are not at the same scale, we also manually annotate the bounding box for the head in these training set images, and bring them to canonical scale. Further, we crop them to $320 \times 240$ such that the center of the shoulder annotations lies at (160 px, 80 px). We do not perform this image normalization at test time. Following the methodology of Felzenszwalb et al.~\cite{felzenszwalb2008discriminatively}, at test time we run our model on images with only one person (351 images of the 1016 test examples). As stated in Section~\ref{sec:model}, the model is run on 6 different input image scales and we then use the joint location with highest confidence across those scales as the final location.

For training the convnet we use Theano~\cite{theano}, which provides a Python-based framework for efficient GPU processing and symbolic differentiation of complex compound functions. To reduce GPU memory usage while training, we cache only 100 mini-batches on the GPU; this allows us to use larger convnet models and keep all training data on a single GPU. As part of this framework, our system has two main threads of execution: 1) a training function which runs on the GPU evaluating the batched-SGD updates, and 2) a data dispatch function which preprocesses the data on the CPU and transfers it on the GPU when thread 1) is finished processing the 100 mini batches. Training each convnet on an NVIDIA TITAN GPU takes 1.9ms per patch (fprop + bprop) = 41min total. We test on a cpu cluster with 5000 nodes. Testing takes: 0.49sec per image (0.94x scale) = 2.8min total.
NMS and spatial model take negligible time.

For testing, because of the shared nature of weights for all windows in each image, we convolve the learned filters with the full image instead of individual windows.  This dramatically reduces the time to perform forward propagation on the full test set.

\subsection{Evaluation}

\begin{figure}[ht]
\begin{tabular}{ccc}
\includegraphics[width=0.3\linewidth]{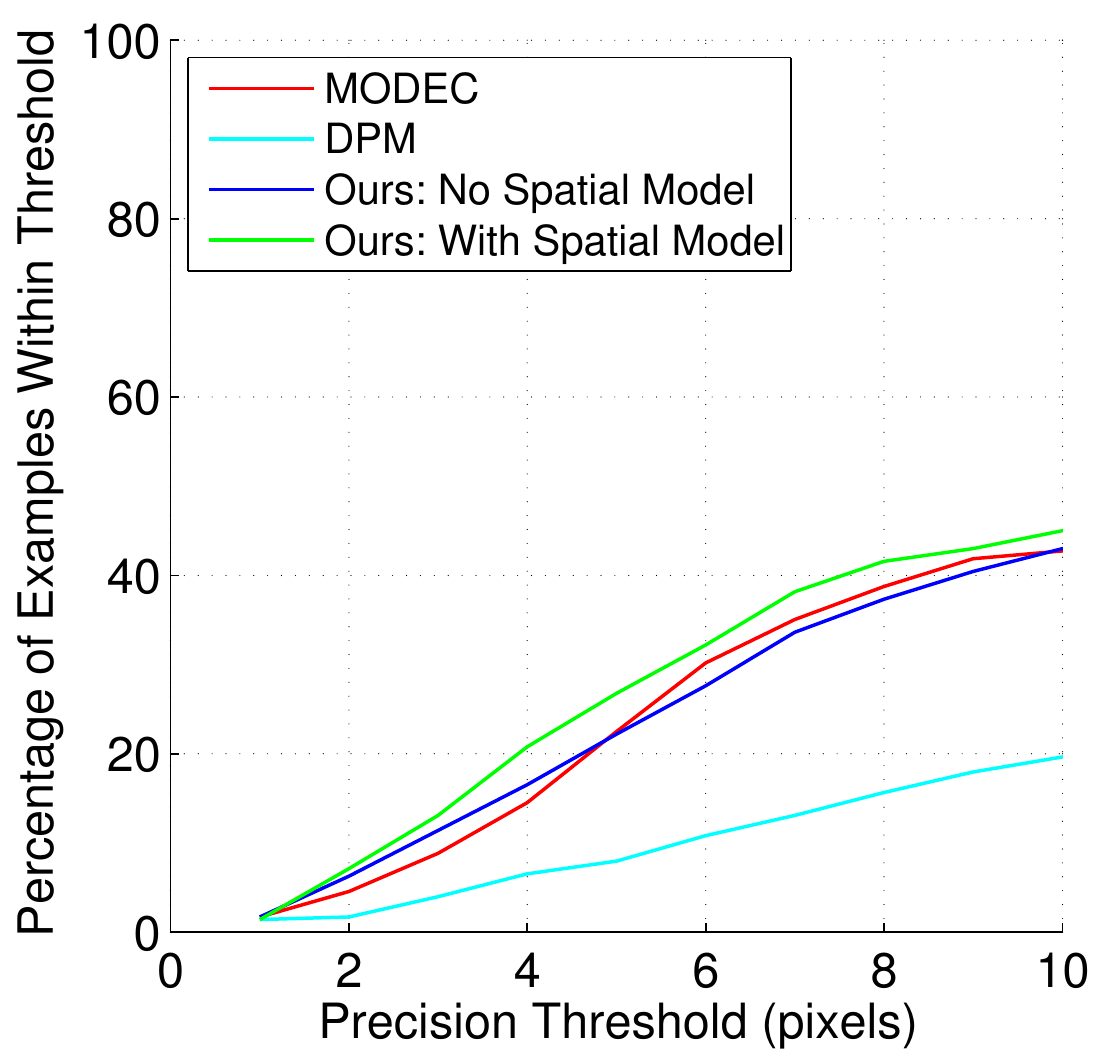} &
\includegraphics[width=0.3\linewidth]{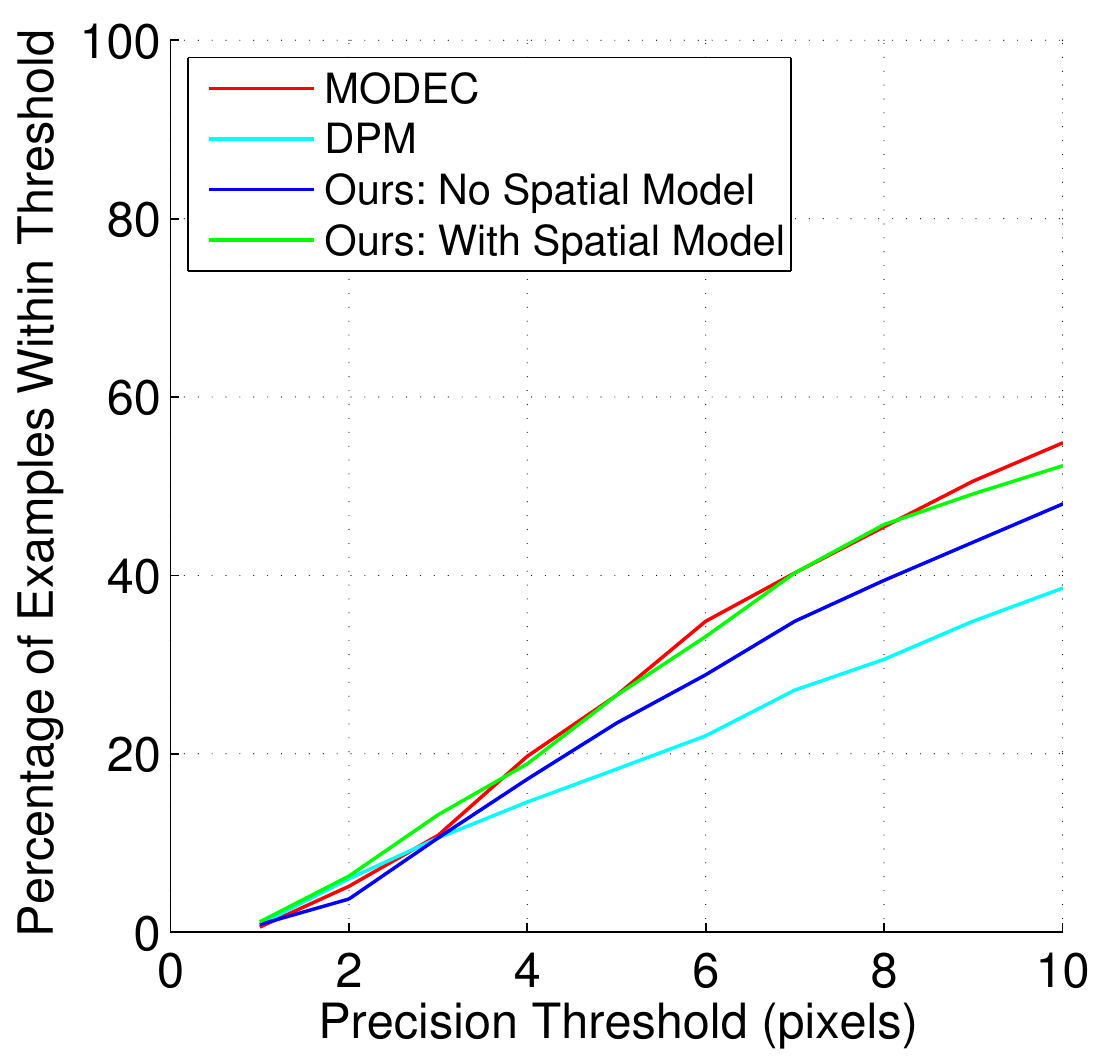} &
\includegraphics[width=0.3\linewidth]{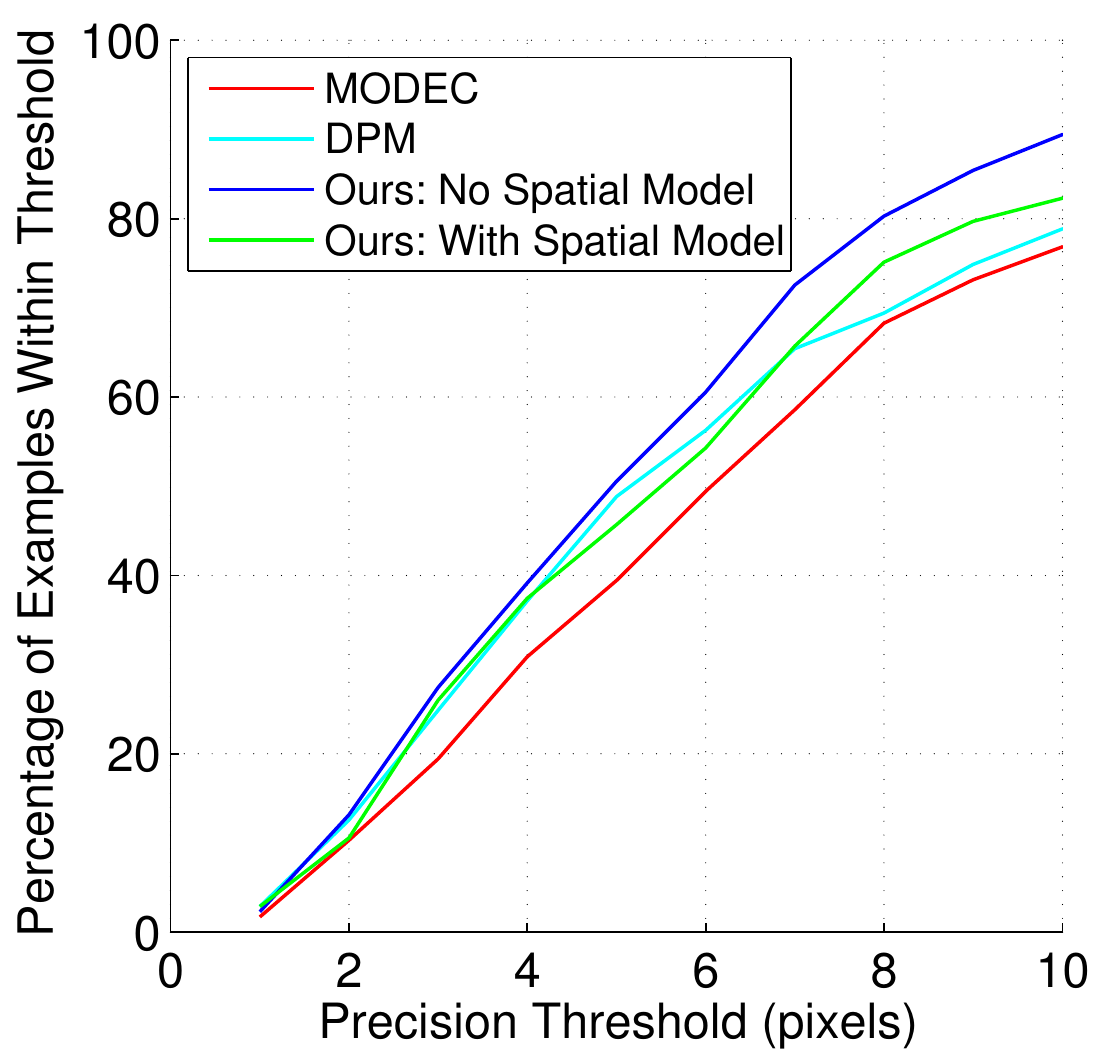} \\
a) Wrist & b) Elbow & c) Shoulder
\end{tabular}
\label{fig:allresults}
\caption{Comparison of Detector Performance on the Test set}
\end{figure}

To evaluate our model on the FLIC dataset we use a measure of accuracy suggested by Sapp et al.~\cite{sapp13cvpr}: for a given joint precision radius we report the percentage of joints in the test set correct within the radius threshold (where distance is defined as 2D Euclidean distance in pixels). In Figure~\ref{fig:allresults} we evaluate this performance measure on the the wrist, elbow and shoulder joints. We also compare our detector to the DPM \cite{felzenszwalb2008discriminatively} and MODEC \cite{sapp13cvpr} architectures.  Note that we use the same subset of 351 images when testing all detectors.


Figure~\ref{fig:allresults} shows that our architecture out-performs or is equal to the MODEC and DPM detectors for all three body parts.  For the wrist and elbow joints our simple spatial model improves joint localization for approximately 5\% of the test set cases (at a 5 pixel threshold), which enables us to outperform all other detectors. However, for the shoulder joint our spatial model actual decreases the joint location accuracy for large thresholds.  This is likely due to the poor performance of the convnet on the elbow.

\begin{figure}[ht]
\begin{tabular}{ccc}
\includegraphics[width=0.3\linewidth]{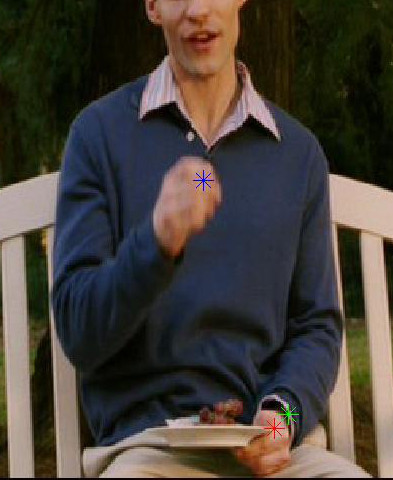} &
\includegraphics[width=0.3\linewidth]{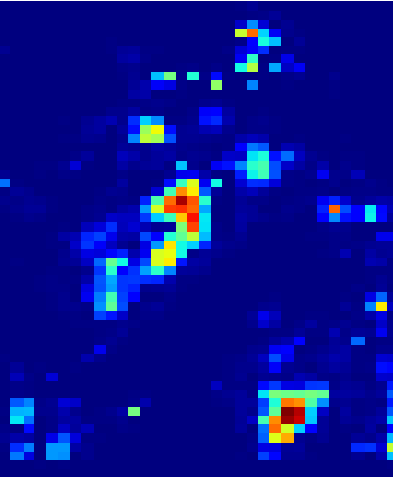} &
\includegraphics[width=0.3\linewidth]{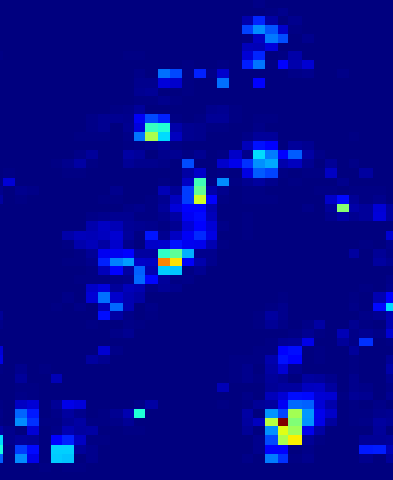} \\
a) RGB and joints & b) distribution before & c) distribution after spatial model.
\end{tabular}
\label{fig:sp_example}
\caption{Impact of Our Spatial Model: Red cross is MODEC, Blue cross is before our Spatial Model, Green cross is after our Spatial Model}      
\end{figure}

As expected, the spatial model cannot improve the joint accuracy of points that are already close to the correct value, however it is never-the-less successful in removing outliers for the wrist and elbow joints.  Figure~\ref{fig:sp_example} is an example where a strong false positive results in an incorrect part location before the spatial model is applied, which is subsequently removed after applying our spatial model.

%

%


\begin{figure}[ht]
\begin{center}
\includegraphics[width=1\linewidth]{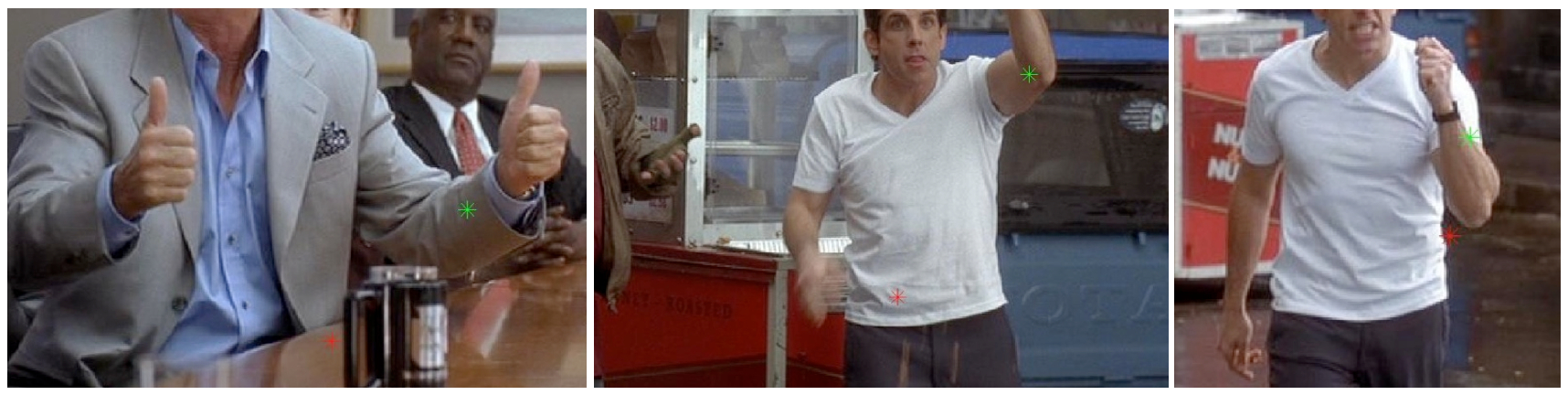}
\end{center}
\label{fig:failure_pix}
\caption{Failure cases: The green cross is our new technique's wrist locator, the red cross is the state-of-the-art CVPR13 MODEC detector \cite{sapp13cvpr} on the FLIC database.}
\end{figure}
\begin{figure}[ht]
\begin{center}
\includegraphics[width=1\linewidth]{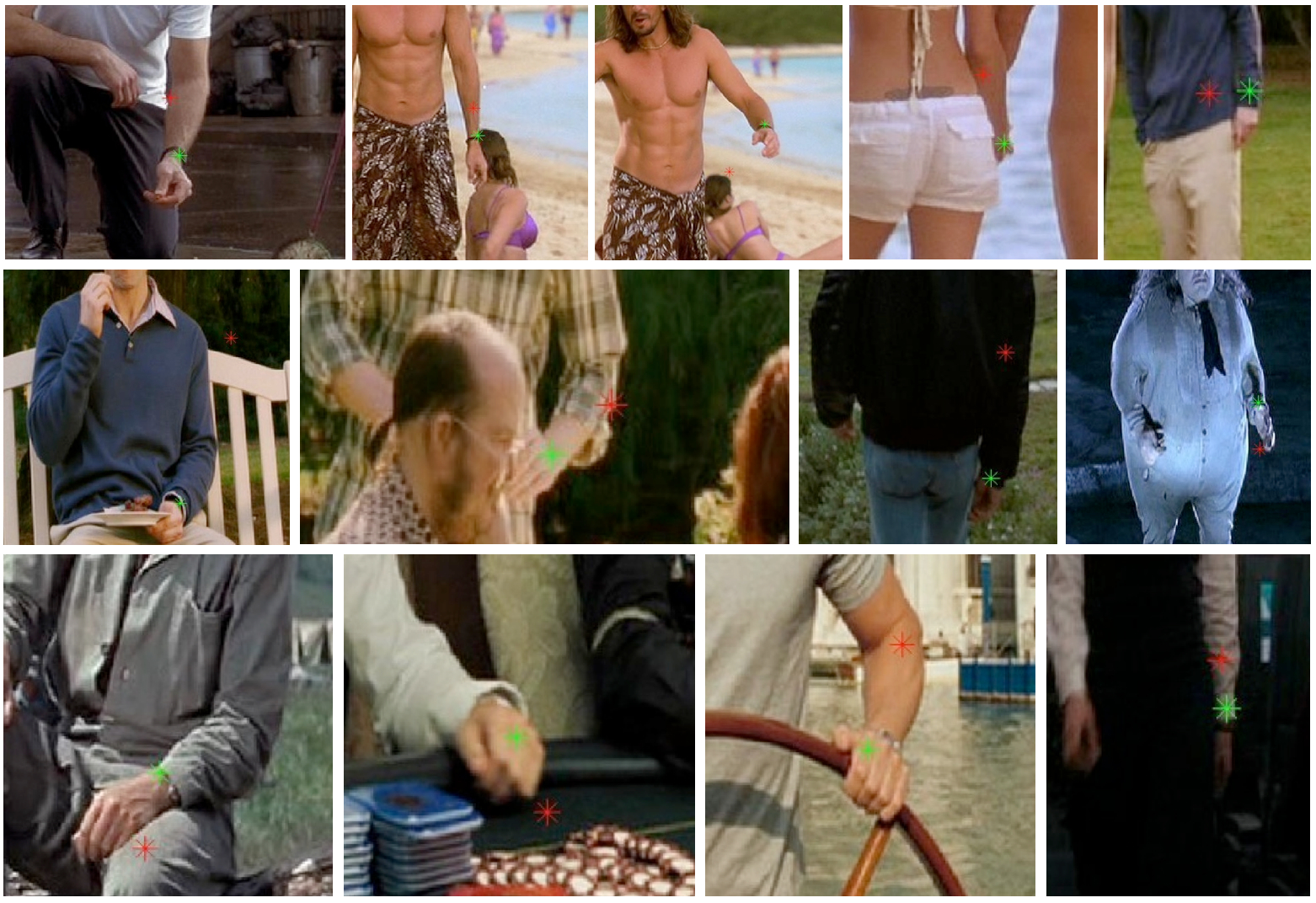}
\end{center}
\label{fig:success_pix}
\caption{Success cases: The green cross is our new technique's wrist locator, the red cross is the state-of-the-art CVPR13 MODEC detector \cite{sapp13cvpr} on the FLIC database.}
\end{figure}

\section{Conclusion}

We have shown successfully how to improve the state-of-the-art on one of the most complex computer vision tasks: unconstrained human pose estimation.  
Convnets are impressive low-level feature detectors, which when combined with a global position prior is able to outperform much more complex and popular models. 
 We explored many different higher level structural models with the aim to further improve the results, but the most generic higher level spatial model achieved the best results.  As mentioned in the introduction, this is counter-intuitive to common belief for human kinematic structures, but it mirrors results in other domains.  For instance in speech recognition, researchers observed, if the learned transition probabilities (higher level structure) are reset to equal probabilities, the recognition performance, now mainly driven by the emission probabilities does not reduce significantly \cite{MorganPrivateHMM}.  Other domains are discussed in more detail by \cite{lucchi2011spatial}.

We expect to obtain further improvement by enlarging the training set with a new pose-based warping technique that we are currently investigating.  Furthermore, we are also currently experimenting with multi-resolution input representations, that take a larger spatial context into account.


\section{Acknowledgements}

This research was funded in part by the Office of Naval Research ONR Award N000141210327 and by a Google award.

{\small
\bibliographystyle{ieee}
\bibliography{biblio}

\begin{thebibliography}{10}\itemsep=-1pt

\bibitem{agarwal2006recovering}
A.~Agarwal, B.~Triggs, I.~Rhone-Alpes, and F.~Montbonnot.
\newblock {Recovering 3D human pose from monocular images}.
\newblock {\em IEEE Transactions on Pattern Analysis and Machine Intelligence},
  28(1):44--58, 2006.

\bibitem{andriluka2009pictorial}
M.~Andriluka, S.~Roth, and B.~Schiele.
\newblock {Pictorial structures revisited: People detection and articulated
  pose estimation}.
\newblock In {\em CVPR}, 2009.

\bibitem{andriluka2010monocular}
M.~Andriluka, S.~Roth, and B.~Schiele.
\newblock Monocular 3d pose estimation and tracking by detection.
\newblock In {\em Computer Vision and Pattern Recognition (CVPR), 2010 IEEE
  Conference on}, pages 623--630. IEEE, 2010.

\bibitem{anguelov2005scape}
D.~Anguelov, P.~Srinivasan, D.~Koller, S.~Thrun, J.~Rodgers, and J.~Davis.
\newblock Scape: shape completion and animation of people.
\newblock In {\em ACM Transactions on Graphics (TOG)}, volume~24, pages
  408--416. ACM, 2005.

\bibitem{athitsos2004boostmap}
V.~Athitsos, J.~Alon, S.~Sclaroff, and G.~Kollios.
\newblock {Boostmap: A method for efficient approximate similarity rankings}.
\newblock {\em CVPR}, 2004.

\bibitem{Bengio2012}
Y.~Bengio, A.~C. Courville, and P.~Vincent.
\newblock Representation learning: A review and new perspectives.
\newblock Technical report, University of Montreal, 2012.

\bibitem{theano}
J.~Bergstra, O.~Breuleux, F.~Bastien, P.~Lamblin, R.~Pascanu, G.~Desjardins,
  J.~Turian, D.~Warde-Farley, and Y.~Bengio.
\newblock Theano: a {CPU} and {GPU} math expression compiler.
\newblock In {\em Proceedings of the Python for Scientific Computing Conference
  ({SciPy})}, June 2010.
\newblock Oral Presentation.

\bibitem{Coreg:Patent:2012}
M.~Black, D.~Hirshberg, M.~Loper, E.~Rachlin, and A.~Weiss.
\newblock Co-registration -- simultaneous alignment and modeling of articulated
  {3D} shapes.
\newblock European patent application EP12187467.1 and US Provisional
  Application, Oct. 2012.

\bibitem{PoseletsICCV09}
L.~Bourdev and J.~Malik.
\newblock Poselets: Body part detectors trained using 3d human pose
  annotations.
\newblock In {\em ICCV}, sep 2009.

\bibitem{buehler2009learning}
P.~Buehler, A.~Zisserman, and M.~Everingham.
\newblock {Learning sign language by watching TV (using weakly aligned
  subtitles)}.
\newblock {\em CVPR}, 2009.

\bibitem{dalal2005histograms}
N.~Dalal and B.~Triggs.
\newblock Histograms of oriented gradients for human detection.
\newblock In {\em Computer Vision and Pattern Recognition, 2005. CVPR 2005.
  IEEE Computer Society Conference on}, volume~1, pages 886--893. IEEE, 2005.

\bibitem{erol2007vision}
A.~Erol, G.~Bebis, M.~Nicolescu, R.~D. Boyle, and X.~Twombly.
\newblock Vision-based hand pose estimation: A review.
\newblock {\em Computer Vision and Image Understanding}, 108(1):52--73, 2007.

\bibitem{FarabetCouprieNajmanLeCun2012}
C.~Farabet, C.~Couprie, L.~Najman, and Y.~LeCun.
\newblock Scene parsing with multiscale feature learning, purity trees, and
  optimal covers.
\newblock In {\em ICML}, 2012.

\bibitem{farhadi2007transfer}
A.~Farhadi, D.~Forsyth, and R.~White.
\newblock {Transfer Learning in Sign language}.
\newblock In {\em CVPR}, 2007.

\bibitem{felzenszwalb2008discriminatively}
P.~Felzenszwalb, D.~McAllester, and D.~Ramanan.
\newblock {A discriminatively trained, multiscale, deformable part model}.
\newblock In {\em CVPR}, 2008.

\bibitem{Felzenszwalb2010PAMI}
P.~F. Felzenszwalb, R.~B. Girshick, D.~McAllester, and D.~Ramanan.
\newblock Object detection with discriminatively trained part-based models.
\newblock {\em PAMI'10}.

\bibitem{Ferrari2009}
V.~Ferrari, M.~Marin-Jimenez, and A.~Zisserman.
\newblock Pose search: {R}etrieving people using their pose.
\newblock In {\em CVPR}, 2009.

\bibitem{glorot2011deep}
X.~Glorot, A.~Bordes, and Y.~Bengio.
\newblock Deep sparse rectifier networks.
\newblock In {\em Proceedings of the 14th International Conference on
  Artificial Intelligence and Statistics. JMLR W\&CP Volume}, volume~15, pages
  315--323, 2011.

\bibitem{Grauman2003}
K.~Grauman, G.~Shakhnarovich, and T.~Darrell.
\newblock Inferring 3d structure with a statistical image-based shape model.
\newblock In {\em ICCV}, pages 641--648, 2003.

\bibitem{HasStoSunRosSei09}
N.~Hasler, C.~Stoll, M.~Sunkel, B.~Rosenhahn, and H.-P. Seidel.
\newblock A statistical model of human pose and body shape.
\newblock In P.~Dutr{'e} and M.~Stamminger, editors, {\em Computer Graphics
  Forum (Proc. Eurographics 2008)}, volume~2, Munich, Germany, Mar. 2009.

\bibitem{hinton2012improving}
G.~E. Hinton, N.~Srivastava, A.~Krizhevsky, I.~Sutskever, and R.~R.
  Salakhutdinov.
\newblock Improving neural networks by preventing co-adaptation of feature
  detectors.
\newblock {\em arXiv preprint arXiv:1207.0580}, 2012.

\bibitem{yann_lcn_cite}
K.~Jarrett, K.~Kavukcuoglu, M.~Ranzato, and Y.~LeCun.
\newblock What is the best multi-stage architecture for object recognition?
\newblock In {\em Computer Vision, 2009 IEEE 12th International Conference on},
  pages 2146--2153, Sept 2009.

\bibitem{krizhevsky2012imagenet}
A.~Krizhevsky, I.~Sutskever, and G.~Hinton.
\newblock Imagenet classification with deep convolutional neural networks.
\newblock In {\em Advances in Neural Information Processing Systems 25}, pages
  1106--1114, 2012.

\bibitem{LeCun1998}
Y.~LeCun, L.~Bottou, Y.~Bengio, and P.~Haffner.
\newblock Gradient-based learning applied to document recognition.
\newblock {\em Proc. IEEE}, 86(11):2278--2324, 1998.

\bibitem{lowe1999object}
D.~G. Lowe.
\newblock Object recognition from local scale-invariant features.
\newblock In {\em Computer vision, 1999. The proceedings of the seventh IEEE
  international conference on}, volume~2, pages 1150--1157. Ieee, 1999.

\bibitem{lucchi2011spatial}
A.~Lucchi, Y.~Li, X.~Boix, K.~Smith, and P.~Fua.
\newblock Are spatial and global constraints really necessary for segmentation?
\newblock In {\em Computer Vision (ICCV), 2011 IEEE International Conference
  on}, pages 9--16. IEEE, 2011.

\bibitem{MorganPrivateHMM}
N.~Morgan.
\newblock personal communication.

\bibitem{mori2002estimating}
G.~Mori and J.~Malik.
\newblock {Estimating human body configurations using shape context matching}.
\newblock {\em ECCV}, 2002.

\bibitem{Neubeck:2006:ENS:1170749.1172615}
A.~Neubeck and L.~Van~Gool.
\newblock Efficient non-maximum suppression.
\newblock In {\em Proceedings of the 18th International Conference on Pattern
  Recognition - Volume 03}, ICPR '06, pages 850--855, Washington, DC, USA,
  2006. IEEE Computer Society.

\bibitem{nowlan1995convolutional}
S.~J. Nowlan and J.~C. Platt.
\newblock A convolutional neural network hand tracker.
\newblock {\em Advances in Neural Information Processing Systems}, pages
  901--908, 1995.

\bibitem{osadchy2007synergistic}
M.~Osadchy, Y.~L. Cun, and M.~L. Miller.
\newblock Synergistic face detection and pose estimation with energy-based
  models.
\newblock {\em The Journal of Machine Learning Research}, 8:1197--1215, 2007.

\bibitem{pinto2008real}
N.~Pinto, D.~D. Cox, and J.~J. DiCarlo.
\newblock Why is real-world visual object recognition hard?
\newblock {\em PLoS computational biology}, 4(1):e27, 2008.

\bibitem{pishchulin12cvpr}
L.~Pishchulin, A.~Jain, M.~Andriluka, T.~Thormaehlen, and B.~Schiele.
\newblock Articulated people detection and pose estimation: Reshaping the
  future.
\newblock In {\em CVPR'12}.

\bibitem{pishchulin11bmvc}
L.~Pishchulin, A.~Jain, C.~Wojek, T.~Thormaehlen, and B.~Schiele.
\newblock In good shape: Robust people detection based on appearance and shape.
\newblock In {\em BMVC'11}.

\bibitem{poppe2007vision}
R.~Poppe.
\newblock {Vision-based human motion analysis: An overview}.
\newblock {\em Computer Vision and Image Understanding}, 108(1-2):4--18, 2007.

\bibitem{ramanan2005strike}
D.~Ramanan, D.~Forsyth, and A.~Zisserman.
\newblock {Strike a pose: Tracking people by finding stylized poses}.
\newblock In {\em CVPR}, 2005.

\bibitem{Sapp2010}
B.~Sapp, C.~Jordan, and B.Taskar.
\newblock Adaptive pose priors for pictorial structures.
\newblock In {\em CVPR}, 2010.

\bibitem{sapp13cvpr}
B.~Sapp and B.~Taskar.
\newblock Multimodal decomposable models for human pose estimation.
\newblock In {\em CVPR'13}.

\bibitem{Shakhnarovich2003}
G.~Shakhnarovich, P.~Viola, and T.~Darrell.
\newblock Fast pose estimation with parameter-sensitive hashing.
\newblock In {\em ICCV}, pages 750--759, 2003.

\bibitem{shotton2013real}
J.~Shotton, T.~Sharp, A.~Kipman, A.~Fitzgibbon, M.~Finocchio, A.~Blake,
  M.~Cook, and R.~Moore.
\newblock Real-time human pose recognition in parts from single depth images.
\newblock {\em Communications of the ACM}, 56(1):116--124, 2013.

\bibitem{Sigal2010}
L.~Sigal, A.~Balan, and B.~M. J.
\newblock {HumanEva: Synchronized video and motion capture dataset and baseline
  algorithm for evaluation of articulated human motion}.
\newblock {\em IJCV}, 87(1/2):4--27, 2010.

\bibitem{stoll2011fast}
C.~Stoll, N.~Hasler, J.~Gall, H.~Seidel, and C.~Theobalt.
\newblock Fast articulated motion tracking using a sums of gaussians body
  model.
\newblock In {\em Computer Vision (ICCV), 2011 IEEE International Conference
  on}, pages 951--958. IEEE, 2011.

\bibitem{sutskeverimportance}
I.~Sutskever, J.~Martens, G.~Dahl, and G.~Hinton.
\newblock On the importance of initialization and momentum in deep learning.

\bibitem{taylor2010embedding}
G.~Taylor, R.~Fergus, I.~Spiro, G.~Williams, and C.~Bregler.
\newblock Pose-sensitive embedding by nonlinear {NCA} regression.
\newblock In {\em Advances in Neural Information Processing Systems 23 (NIPS)},
  pages 2280--2288, 2010.

\bibitem{taylor2010tracking}
G.~Taylor, L.~Sigal, D.~Fleet, and G.~Hinton.
\newblock Dynamical binary latent variable models for 3d human pose tracking.
\newblock In {\em Proc.~ of the 23rd IEEE Computer Society Conference on
  Computer Vision and Pattern Recognition (CVPR)}, 2010.

\bibitem{tieleman2012rmsprop}
T.~Tieleman and G.~Hinton.
\newblock Lecture 6.5-rmsprop: Divide the gradient by a running average of its
  recent magnitude.
\newblock {\em COURSERA: Neural Networks for Machine Learning}, 2012.

\bibitem{Turaga2010}
S.~C. Turaga, J.~F. Murray, V.~Jain, F.~Roth, M.~Helmstaedter, K.~Briggman,
  W.~Denk, and H.~S. Seung.
\newblock Convolutional networks can learn to generate affinity graphs for
  image segmentation.
\newblock {\em Neural Computation}, 22:511--538, 2010.

\bibitem{wang2009real}
R.~Y. Wang and J.~Popovi{\'c}.
\newblock Real-time hand-tracking with a color glove.
\newblock In {\em ACM Transactions on Graphics (TOG)}, volume~28, page~63. ACM,
  2009.

\bibitem{wren1997pfinder}
C.~Wren, A.~Azarbayejani, T.~Darrell, and A.~Pentland.
\newblock {Pfinder: Real-time tracking of the human body}.
\newblock {\em IEEE Transactions on Pattern Analysis and Machine Intelligence},
  19(7):780--785, 1997.

\bibitem{yang2011articulated}
Y.~Yang and D.~Ramanan.
\newblock Articulated pose estimation with flexible mixtures-of-parts.
\newblock In {\em Computer Vision and Pattern Recognition (CVPR), 2011 IEEE
  Conference on}, pages 1385--1392. IEEE, 2011.

\bibitem{zeiler2013visualizing}
M.~D. Zeiler and R.~Fergus.
\newblock Visualizing and understanding convolutional neural networks.
\newblock {\em arXiv preprint arXiv:1311.2901}, 2013.

\bibitem{zuffiestimating}
S.~Zuffi, J.~Romero, C.~Schmid, and M.~J. Black.
\newblock Estimating human pose with flowing puppets.

\end{thebibliography}
}

\end{document}